# Virtual Dress Swap Using Landmark Detection

**Odar Zeynal**
Kapital Bank
Department of Consumer Experience
oder.zeynal@gmail.com

**Saber Malekzadeh**
Khazar University
Lumos Extension School of Data
saber.malekzadeh@sru.ac.ir

*Abstract -* **Online shopping has gained popularity recently. This paper addresses one crucial problem of buying dress online, which has not been solved yet. This research tries to implement the idea of clothes swapping with the help of DeepFashion dataset where 6,223 images with eight landmarks each used. Deep Convolutional Neural Network has been built for Landmark detection.**

## I. INTRODUCTION

We are living in the era where internet take indispensable role in our lives and one part of its role is through e-commerce. There is a steady increase in online shopping not only by the amount but by also count of users [1]. Moreover, pandemic situation, which brought lock-downs in many countries, boosted online shopping more since there was no other way to acquire goods and services. Clothes and apparel has become favorite part of online shopping. Customers, without going out of their home, can look, compare and decide what they want to dress through various online shopping websites. Although e-commerce has many advantages such as being easier and cheaper, costing less time and resources for both customers and sellers, it has also disadvantages. One of its disadvantages is that online shopping websites cannot answer the "how will it look on me?" question. Therefore, there are still many people, who are skeptical about e-commerce and prefer physical shopping instead. Furthermore, some customers are not happy after the delivery of their dress [2].

In this research, we try to answer the question mentioned above with the help of DeepFake. Object of this research is to create a digital dressing room mirror where simple selfie camera would be enough for customer to digitally dress the clothes and see how it looks. The term "DeepFake" which is referred to a deep learning based technique, originated after a Reddit user named "deepfakes" in 2017 who has developed a machine-learning algorithm that helped him to transpose faces. Main machine learning methods is used to create deepfakes are based on deep learning and involve training generative neural network architectures such as autoencoders or generative adversarial networks. Face manipulation is most used field of deepfake. Since 2017, development in DeepFakes has been quite overwhelming that nowadays we can produce very realistic fake photos and videos with very small amount data. For example, DeepFake model developed by Samsung AI can fabricate a video from single picture [4]. However, since DeepFakes mainly focused on human face manipulation, people concentrated on replacing faces more than replacing other things such as dresses.

## II. RELATED WORK

Although this is new idea and nothing has been done so far for this matter, it would be useful to get expertise from face manipulation algorithms and implement that knowledge to clothes. Therefore, related work is divided to two parts:

1) Object detection and Landmark Detection for Clothes
2) Face Swap Algorithm

### Object detection and Landmark Detection for Clothes

Different networks has been built on the dependency of tasks, such as clothing recognition, recommendation, retrieval and fashion landmark localization [5]. For example, in 2015 Kiapour and others developed "Where to Buy It" (WTBI) network, which extracted different dress, pictures from streets and referenced to online shopping websites [6]. In the street picture side, they categorise and make a bounding box around the item in the picture. However, they do not label pictures from the shops and rely on algorithm to find and retrieve information. Similarly, J.Huang and others developed Dual Attribute-Aware Ranking Network (DARN) to solve the same issue with more focus on photos taken under uncontrolled conditions like lighting, pose, background etc. [7]. Liu and et. al. came up with largest clothing dataset named DeepFashion which is richly annotated [8]. In DeepFashion dataset, each image is labeled with 50 categories, 1000 descriptive attributes and clothing landmarks. They proposed novel deep model, named FashionNet which simultaneously predicts landmarks and attributes.

### Face Swap Algorithms

Although various researches has been made on DeepFake, most of them, such as DeepFaceLap [9], DFaker [10], DeepFake_tf (tensorflow based deepfakes) [11] use similar approaches which can be categorized in few steps. Face Detection – first of all network should detect face or faces in the picture with bounding boxes. Facial Landmarks – network should understand where facial points such as lips, eyes, nose etc. are located. Face alignment – which makes standardization faces from different poses. Face segmentation – masks aligned face. After these steps, source and destination faces are taken to Autoencoder where they both are encoded with same encoder, but destination face is encoded with source image encoder.

## III. DATASET

In this paper, DeepFashion[8] dataset has been used. DeepFashion dataset includes four sub datasets: Category and Attribute Prediction Benchmark, In-shop Clothes Retrieval Benchmark, Consumer-to-shop Clothes Retrieval Benchmark and Fashion Landmark Detection Benchmark. Landmark Detection Benchmark dataset contains 123,016 distinct images, which falls into three main categories: (upper body, lower body and full body) and other sub-categories such as pose and visibility statuses. In full body picture, there are 8 landmark annotations which coordinates body parts accordingly. For full-body clothes, landmark annotations are listed in the order of left collar, right collar, left sleeve, right sleeve, left waistline, right waistline, left hem, and right hem. From Landmark Detection Benchmark dataset 6,223 distinct picture used in this research with condition of only full-body pictures and visibility of each landmark.

## IV. METHOD

Methodology of this research can be divided into two parts: Landmark Detection and Clothes Swap.

**Landmark Detection**

Landmark detection is performed on Landmark Detection Benchmark dataset mentioned above. Each picture in the dataset is resized from its original size to 100x100x3. Then their landmark positions are adjusted to the new size. The model took image dimensions (100x100x3) and returns an array of shape 16 (8 landmark keypoints * 2 dimensions) as output. The dataset is divided into train (85%) and test (15%). The model starts with Batch Normalisation layer with inputs and continues with five 2D convolutional layers. Mean squared error used as a loss function. Whole structure of the model can be seen in Figure1. In total, we had 1,577,260 neurons trained with batch size 20 for 50 epochs. As a result, mean squared error declined to 126.96 on the validation data. Model tested on test dataset and independent data. Some samples of experiments are in Figure 2.

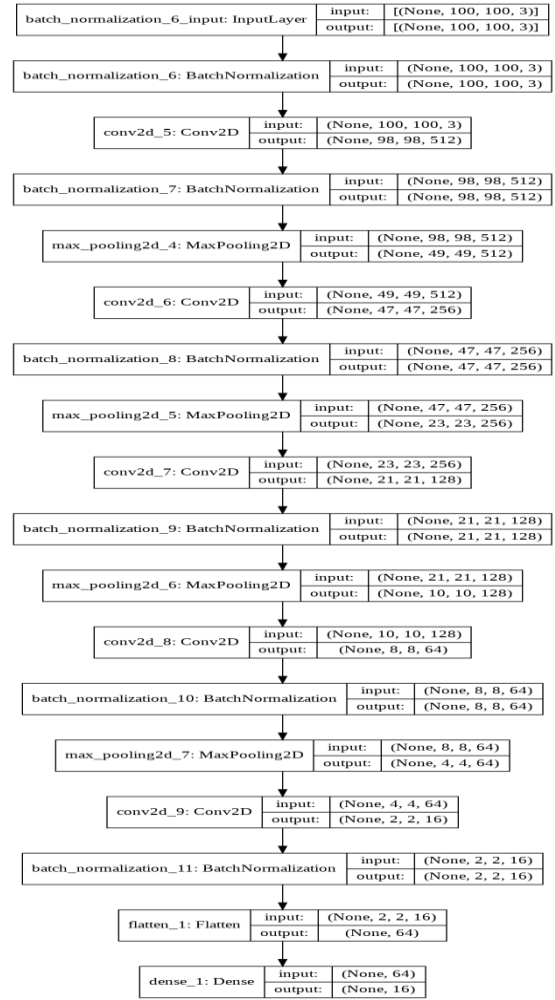

Figure 1 – Structure of the model

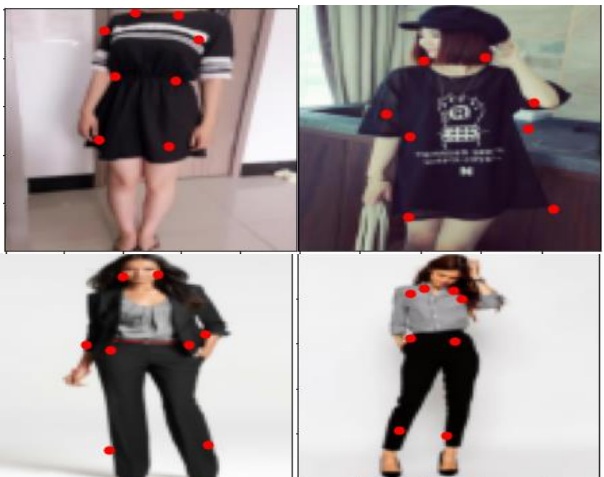

Figure 2 – Sample pictures from experiments

**Clothes Swap**

Once we have coordinates of landmarks, next step is drawing contours and cropping the area inside contours (where clothes locate) from the source picture. However, before cropping we must be sure that our landmark position are in the correct order, which means that our landmark positions must be sorted by clockwise order (or counterclockwise). After cropping the clothes from source image, we need to resize it according to the clothes area of destination image. Finally, we can paste our cropped image on the destination image. In page 6, there is a swap example made by the algorithm mentioned above.

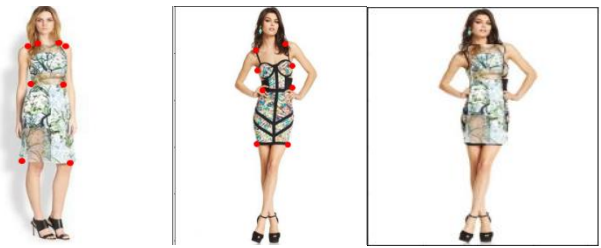

Figure 3 – Swap Example

## V. CONCLUSION

In this paper, DeepFashion dataset used for Landmark detection model and with results of the model clothes-swapping algorithm developed. The model can more or less detect landmarks, but distance between detected landmarks and real landmarks makes swapping difficult. Moreover, swapping with eight landmarks makes the algorithm weak against different poses. Since this is new idea, it can be considered as first try. As a result, few things need to be improved. Firstly, we need more landmarks on a single picture. In this dataset, we had eight landmark for each picture, which makes swapping hard. Secondly, we need customized and tailored dataset. For example, if our goal is to swap t-shirts, we need a dataset with t-shirts only. Otherwise, it tricks the model and makes detection and landmark annotation very hard. Thirdly, we need a stronger model to have better results on various poses.